\documentclass{article}
\usepackage{spconf,amsmath,graphicx}
\usepackage{url}

\title{The Open Brands Dataset: Unified brand detection and recognition at scale}
\name{Xuan Jin, Wei Su, Rong Zhang, Yuan He, Hui Xue}
\address{Alibaba Group}

\begin{document}
\ninept
\maketitle
\begin{abstract}
Intellectual property protection(IPP) have received more and more attention recently due to the development of the global e-commerce platforms. brand recognition plays a significant role in IPP. Recent studies for brand recognition and detection are based on small-scale datasets that are not comprehensive enough when exploring emerging deep learning techniques. Moreover, it is challenging to evaluate the true performance of brand detection methods in realistic and open scenes. In order to tackle these problems, we first define the special issues of brand detection and recognition compared with generic object detection. Second, a novel brands benchmark called "Open Brands" is established. The dataset contains 1,437,812 images which have brands and 50,000 images without any brand. The part with brands in Open Brands  contains 3,113,828 instances annotated in 3 dimensions: 4 types, 559 brands and 1216 logos. To the best of our knowledge, it is the largest dataset for brand detection and recognition with rich annotations. We provide in-depth comprehensive statistics about the dataset, validate the quality of the annotations and study how the performance of many modern models evolves with an increasing amount of training data. Third, we design a network called "Brand Net" to handle brand recognition. Brand Net gets state-of-art mAP on Open Brand compared with existing detection methods.
\end{abstract}
%

\section{Introduction}
\label{sec:intro}

Intellectual Property Protection(IPP) offers protection for inventions, literary and artistic works, symbols, names, and images created by the mind. In e-commerce, IP can be the design and brands of product, and images/videos showed in online shop in e-commerce platforms. brand recognition is a significant part of IP protection. It will protect the entrepreneurs’ and business owners’ hard-earned creations and ideas from unfair competitions. brand recognition is a task that not only aims to recognize the brands but also needs to find the locations of brands.

With the advent of deep convolution network, object detection has witnessed a quantum leap in performance. The current detectors can be divided into two categories: (1) the two-stage approach, including R-CNN\cite{girshick2016region_RCNN}, Fast R-CNN\cite{girshick2015fast_Fast_RCNN}, Faster R-CNN\cite{ren2015faster_Faster_RCNN}, R-FCN\cite{dai2016r_RFCN} and (2) the one-stage approach, including SSD\cite{liu2016ssd_SSD}, YOLO[v1-v3]\cite{redmon2016you_YOLO}\cite{redmon2017yolo9000_YOLO9000}\cite{redmon2018yolov3_YOLOV3}, RetinaNet\cite{lin2018focal_RetinaNet}. In the two-stage approach, a sparse set of candidate object boxes is first generated, and then they are further classified and regressed. The two-stage methods have been achieving top performances on several challenging benchmarks, including Pascal VOC\cite{everingham2010pascal_PASCAL_VOC}, MSCOCO\cite{lin2014microsoft_MSCOCO} and OpenImages\cite{kuznetsova2018open_OpenImage}. 

\begin{figure}
\begin{center}
   \includegraphics[width=0.9\linewidth]{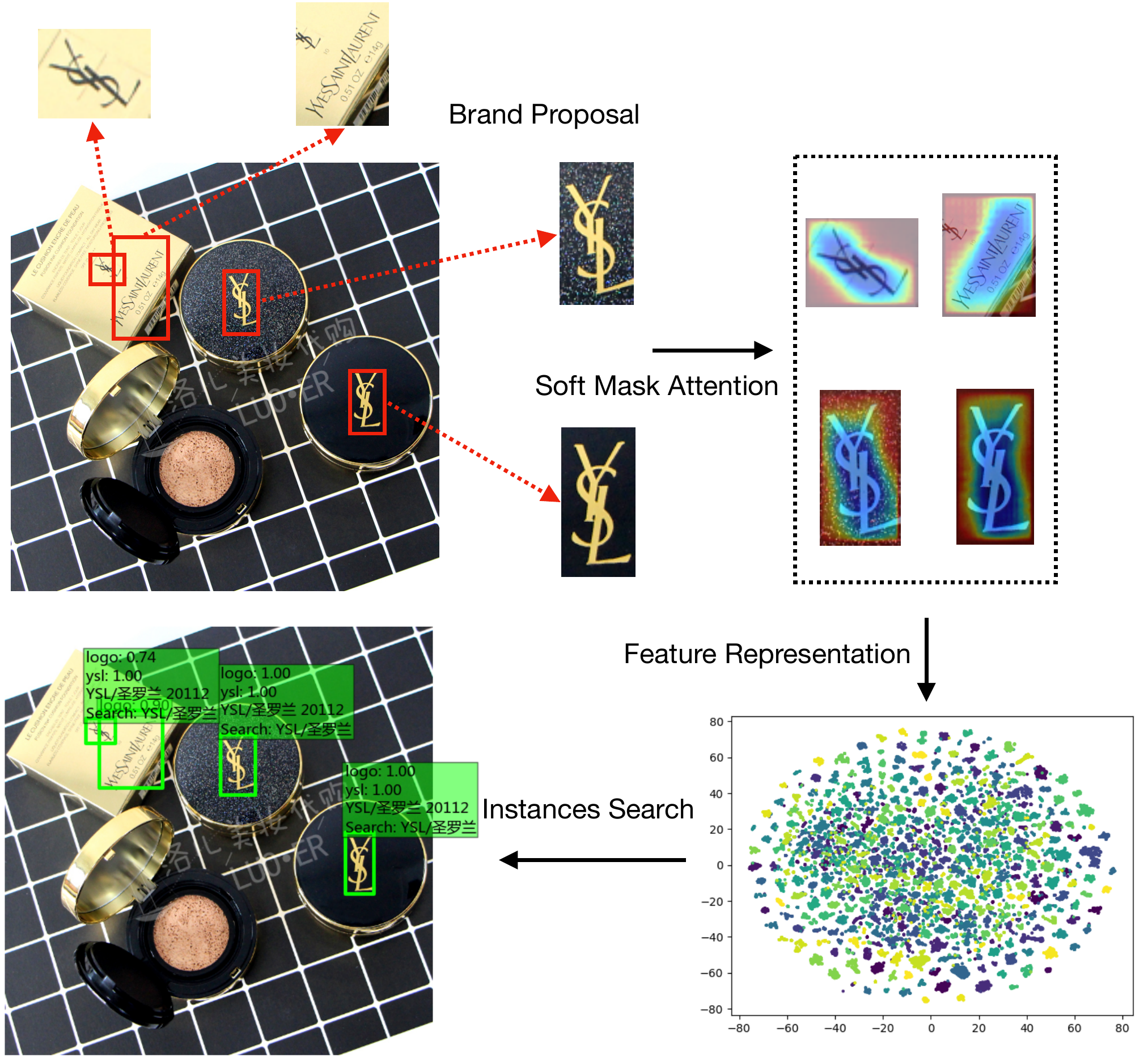}
\end{center}
   \vspace{-5mm}
   \caption{The Framework of Brand Net: Brands proposal, Soft Mask Attention (SMA), Feature Representation and Instance Search.}
\label{fig:Framework}
\end{figure}

brand detection is a special case of generic object detection in computer vision, which is challenging in the following aspects: tiny object detection, large-scale object detection, scalar network, similar brands and instances diversity.

\begin{figure}
\begin{center}
   \includegraphics[width=0.9\linewidth]{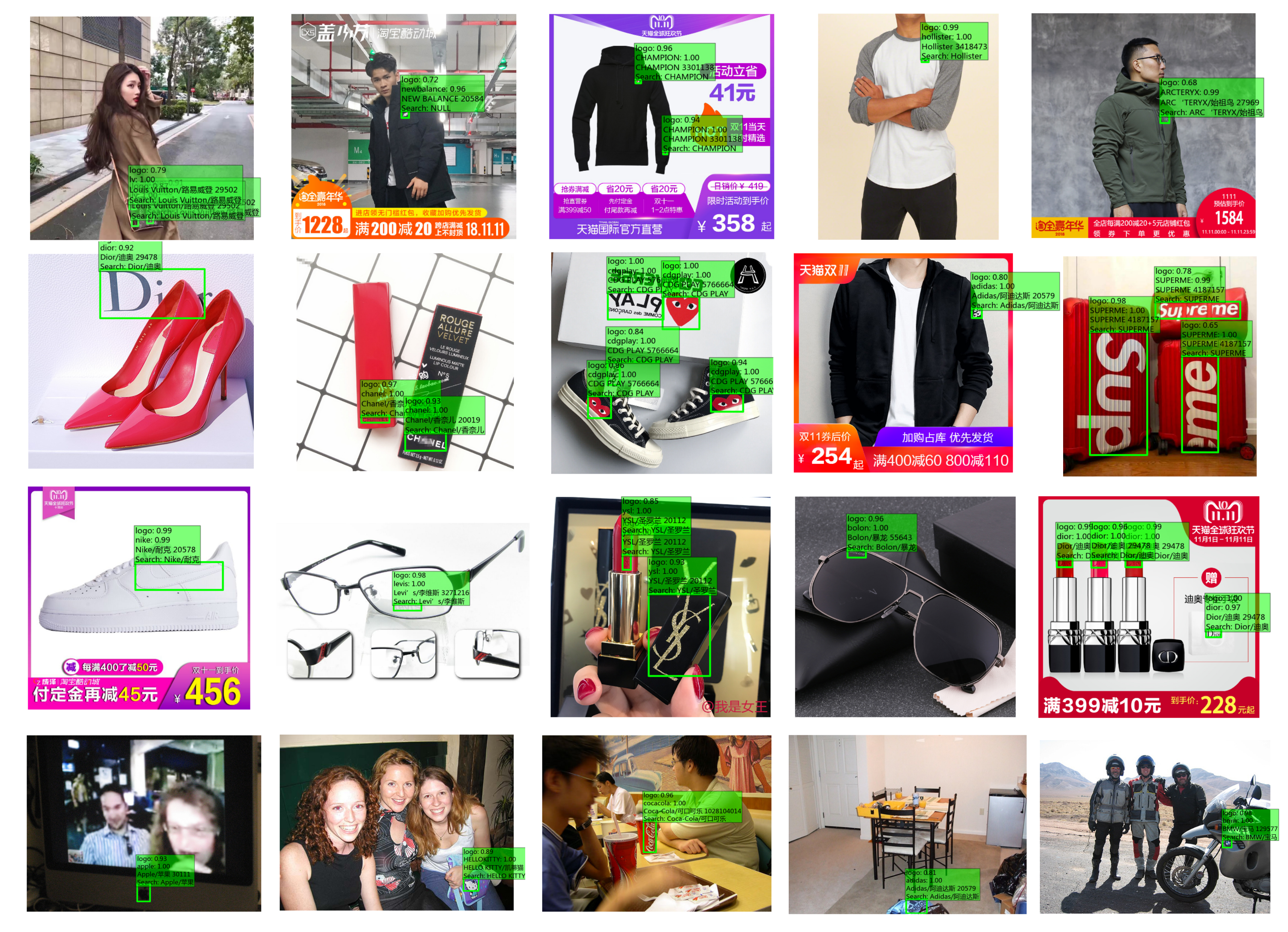}
\end{center}
   \vspace{-5mm}
   \caption{ Diversity: The first row shows samples of tiny brands. The second row shows samples of occlusion. The third row shows samples of brands in low contrast. The forth row shows samples in different scenes: indoor and outdoor. Results in green is provided by Brand Net.}
\label{fig:Diversity}
\end{figure}

{\bf Tiny Objects:} Though tremendous efforts have been made in object detection, one of the remaining open challenges is tiny objects detection due to its limited resolution and information. The object area ratio is much lower in Open Brands than in generic object datasets. As shown in Tab.\ref{tab:Open-Brand and Generic Detection}, the average scale of instances in Open Brands is nearly 17 times smaller than that in Pascal VOC and 8 times smaller than that in MSCOCO and OpenImages. The first row in Fig.\ref{fig:Diversity} gives us some samples of tiny instances, which is even a tough task for human.

{\bf Large-scale Detection:} Current detection methods requiring class-specific sets of filters for each class have hindered their applications for large number of classes. For example, R-FCN/Deformable-R-FCN\cite{dai2017deformable_DFCN} requires 49/197 position-specific filters for each class. RetinaNet requires 9 filters for each class on each convolution feature map. For the billions of products with millions of brands on the e-commerce database,  architectures would need hundreds of millions of filters to detect the tremendous brands, which makes them extremely slow in practical applications. 

{\bf Scalable Network:} Generally the number of categories is fixed when measured on datasets like PASCAL VOC, MS COCO and OpenImages. Nevertheless, a dynamic extension of the categories is a necessary requirement in brand detection since the brands change dynamically with the market. 

\begin{figure}
\begin{center}
   \includegraphics[width=0.9\linewidth]{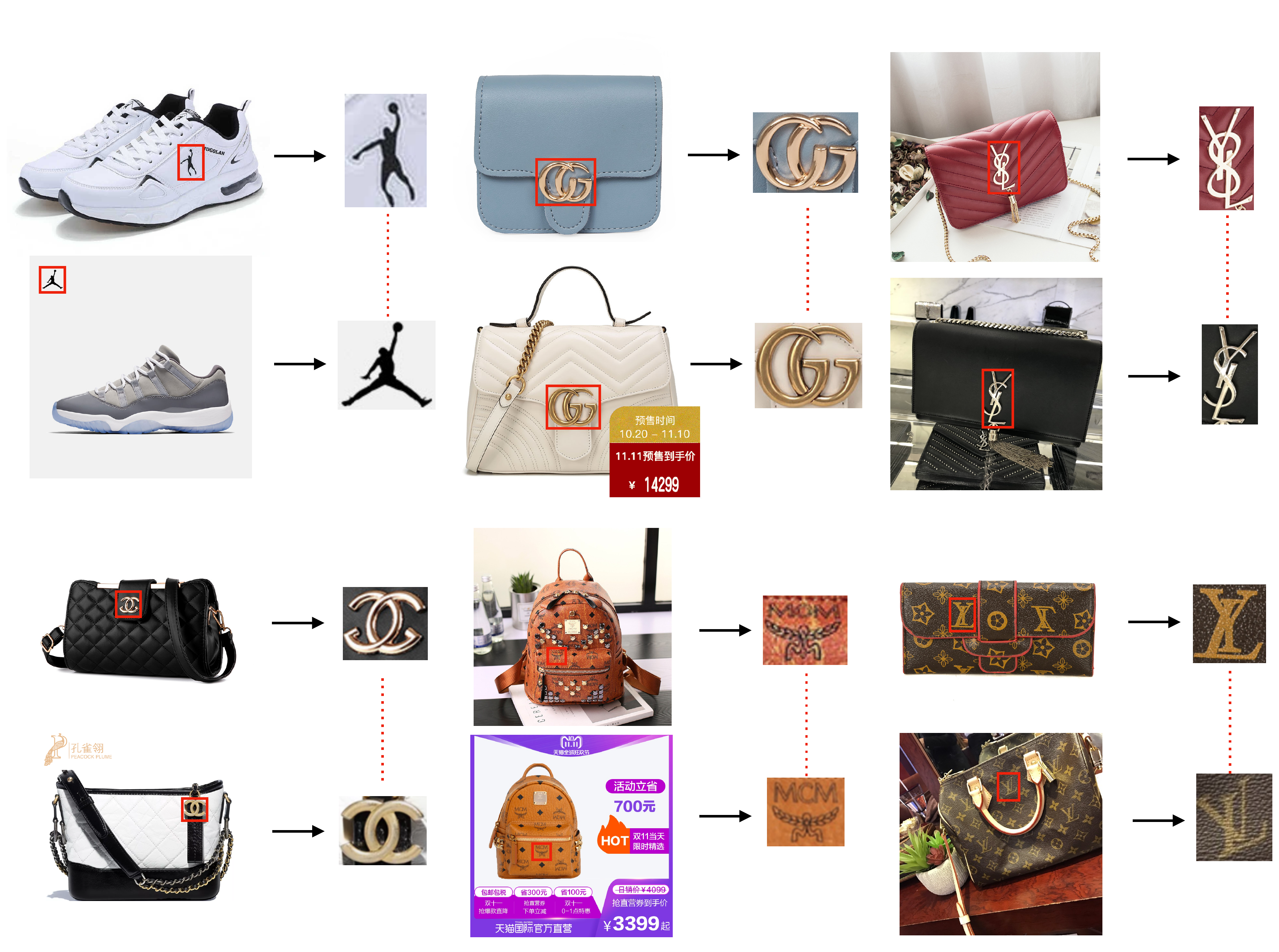}
\end{center}
   \vspace{-5mm}
   \caption{ Some samples of similar brands in Open Brands: "Air Jordan", "Gucci", "Yves Saint Laurent", "Chanel", "MCM", "Louis Vuitton".}
\label{fig:Similar Brand}
\end{figure}

{\bf Similar Brands:} Some brands is so similar that it is difficult for the universal inspection network to tackle. For example, Fig.\ref{fig:Similar Brand} shows some brands that are very similar with the luxury brands. Visually misleading pictures can result in higher error rate of network.

{\bf Diversity:} Diversity is an important factor to judge a dataset. The same brand in real-world may differ a lot caused by different sizes, rotations, transformations, lighting, coloring, and occlusion. A brand could have various appearances on different commodities like clothes, bags, purses, t-shirts and so on. Moreover, the brands appearing on e-commerce platforms are quite different from those appearing in street-scape scenes in Fig.\ref{fig:Diversity}. Last but not the least, the same brand has different logo forms, such as Icons, English words, Chinese and Cartoon pictures.


\begin{table}
\begin{center}
\resizebox{\linewidth}{!}{
\begin{tabular}{l c c c c}
\hline
Datasets & Categories & Images & Mean Instances &  Mean Scale(\%) \\
\hline
PASCAL VOC & 20 & 5717 & 789 & 20.2 \\
MSCOCO & 80 & 118287 & 10750 & 9.65 \\
OpenImages & 600 & 1743042 & 24332 & 9.69 \\
Open Brands & $4|599|1216$ & 1437812 & 2561 & \textbf{1.20} \\
\hline
FlickrLogos\-27 & 27 & 810 & 81 & 19.56 \\
FlickrLogos\-32 & 32 & 2240 & 106 & 9.16 \\
Logo32plus & 32 & 7830 & 338 & 4.51 \\
BelgaLogos & 37 & 1321 & 57 & 0.91 \\
WebLogo\-2M\-test & 194 & 4318 & 41 & 7.69 \\
Logo\-In\-The\-Wild & 1196 & 9393 & 23 & 1.80 \\
SportsLogo & 20 & 1978 & 152 & 9.89 \\
QMUL\-OpenLogo & 352 & 27083 & 88 & 6.09 \\
Logos\-18 & 18 & 8460 & 891 & - \\
Logos\-160 & 160 & 73414 & 816 & - \\
Open Brands & $4|599|1216$ & \textbf{1437812} & \textbf{2561} & 1.20 \\
\hline
\end{tabular}}
\end{center}
\vspace{-5mm}
\caption{Statistic of Open Brands compared with related existing datasets. Mean Scale represents the mean ratio of the instance area to the whole image area. $4|599|1216$ represents 4 types, 599 brands and 1216 logos in Open Brands.}
\label{tab:Open-Brand and Generic Detection}
\end{table}

\section{dataset}
\label{sec:dataset}

\begin{figure}
\begin{center}
   \includegraphics[width=1\linewidth]{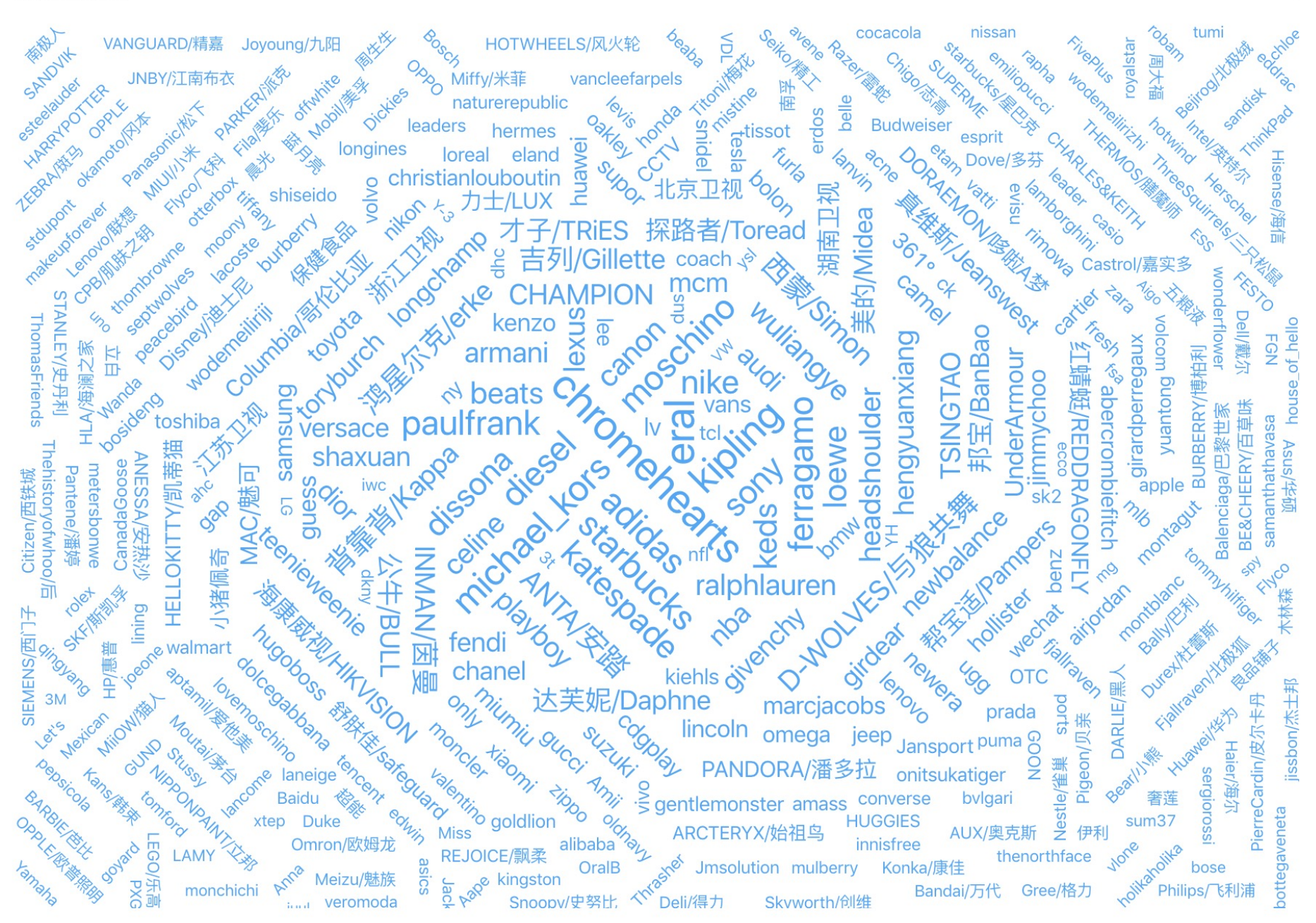}
\end{center}
   \vspace{-5mm}
   \caption{ Most-frequent instance-level brands. Word size is proportional to the brands counts in the training set.
  }
\label{fig:Distribution}
\end{figure}

We contribute Open Brands to the community. The Open Brands contains 1,437,812 images with brands and 50,000 images without brands. The part with brands in Open Brands contains 3,113,828 instances annotated in 3 dimensions: 4 types, 559 brands and 1216 logos. An overview of the most frequent and infrequent brands is shown in Fig.\ref{fig:Distribution}.

{\bf Image Collection:} First of all, we build a tree-like standard library to map a logo to a brand. Then we crawled their product images from several online retail marketplaces: \url{www.taobao.com}, \url{www.tmall.com}, \url{www.1688.com}, \url{www.aliexpress.com} and search engines: \url{www.google.com} and \url{www.baidu.com}. Each brand has more than four viewpoints and $1-4$ logo types. In addition, real scene pictures taken by consumers in the product reviews from online platforms are added. Therefore, the data covers different scenes including both online and street snapshonts.

{\bf Data Cleaning:} We use ResNeXt101\cite{xie2017aggregated} which is trained with hundreds of millions of product images for feature embedding. Images with similar features are duplicated. After the removal of the duplicates, unusable images of low resolutions or poor qualities are removed by human annotators. Finally, 1.4 million images are kept to construct Open Brands.

{\bf Image Annotation:} One of the most time-consuming and costly processes in constructing the Open Brands is annotation of the collected product images. For each product image, a human annotator need to identify the brands, annotate the fixed bounding box of each brand, and then tag it with the corresponding id of types,logos and brands.

{\bf Quality Control:} In order to ensure the quality of the annotations, a task will be dispatched to three human annotators. The annotation label with the highest score which is computed as the intersection-over-union(IoU) is adopted. 

{\bf Comparison to Other Datasets:} Tab.\ref{tab:Open-Brand and Generic Detection} gives statistics of Open Brands compared with existing generic object detection datasets and logo datasets: FlickrLogos-27\cite{kalantidis2011scalable_FlickrLogos27},FlickrLogos-32\cite{romberg2011scalable_FlickrLogos32}, Logo32plus\cite{bianco2017deep_Logo32plus}, BelgaLogos\cite{joly2009logo_BelgaLogos}, WebLogo-2M\cite{su2018weblogo_WebLogo-2M}, Logo-In-The-Wild\cite{tuzko2017open_Logo-In-The-Wild}, SportsLogo\cite{liao2017mutual_SportsLogo}, QMUL-OpenLogo\cite{su2018open_OpenLogo}. Logos-18 and Logos-160 come from Logo-Net\cite{hoi2015logo_Logo-Net}. But as Logo-Net is not public, we can not calculate the mean scale of brands. Above all, Open Brands offers large-scale images and high quality brands instance-aware annotations with localizations and categories. Brand recognition on Open Brands is very challenging due to the tiny instances, large-scale categories, dense diversity and the complicated types of the brands.

\begin{figure}
\begin{center}
   \includegraphics[width=0.9\linewidth]{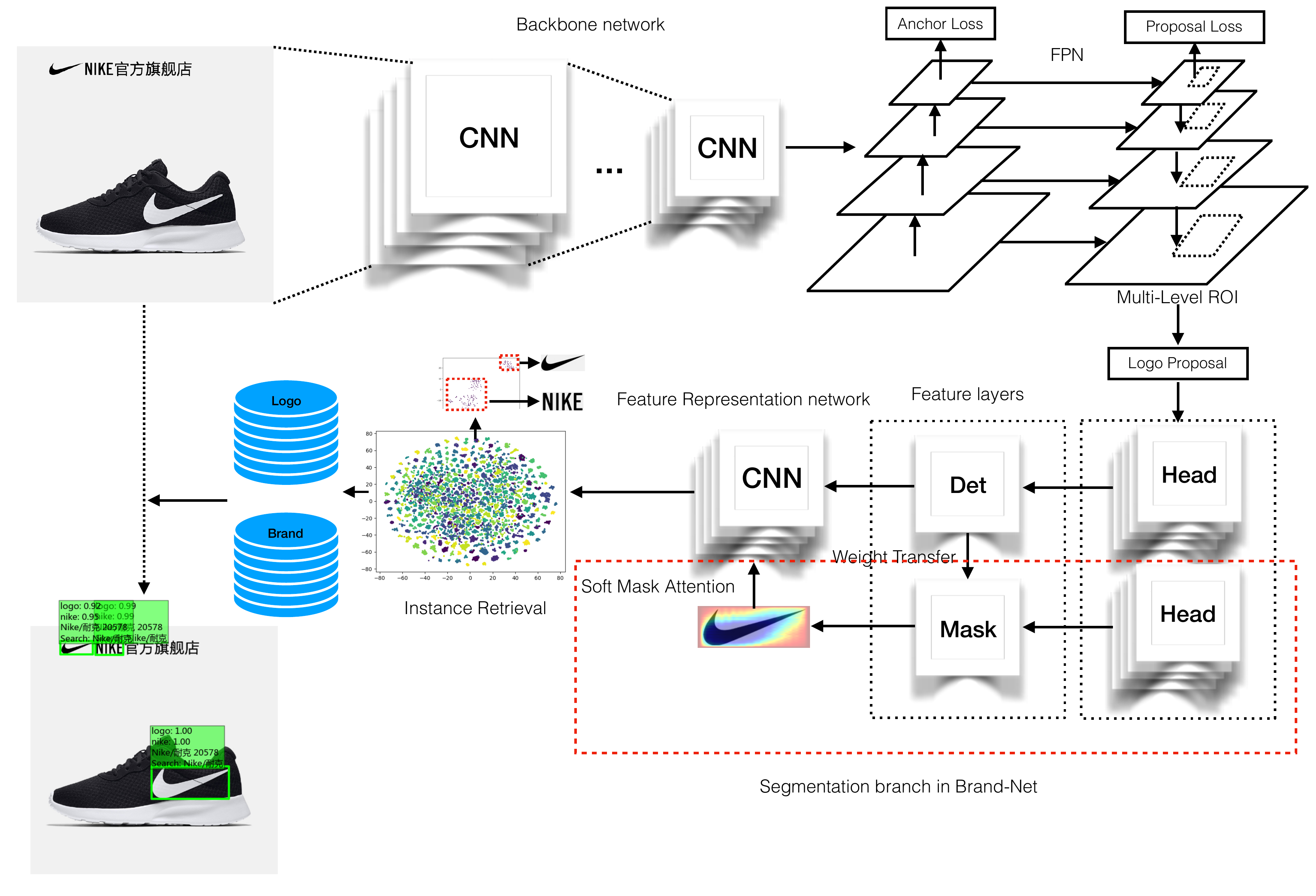}
\end{center}
   \vspace{-5mm}
   \caption{Framework of Brand Net. Soft mask attention which is highlighted in red box is used to improve performance.}
\label{fig: Brand-Net}
\end{figure}

To tackle the problems of brand recognition on Open Brands, we propose a novel network, Brand Net. \ref{fig: Brand-Net} shows the framework of the Brand Net which joints class-agnostic detection, brands feature representation and instances retrieval. Soft mask attention is used for better performance.

\begin{figure}
\begin{center}
   \includegraphics[width=0.9\linewidth]{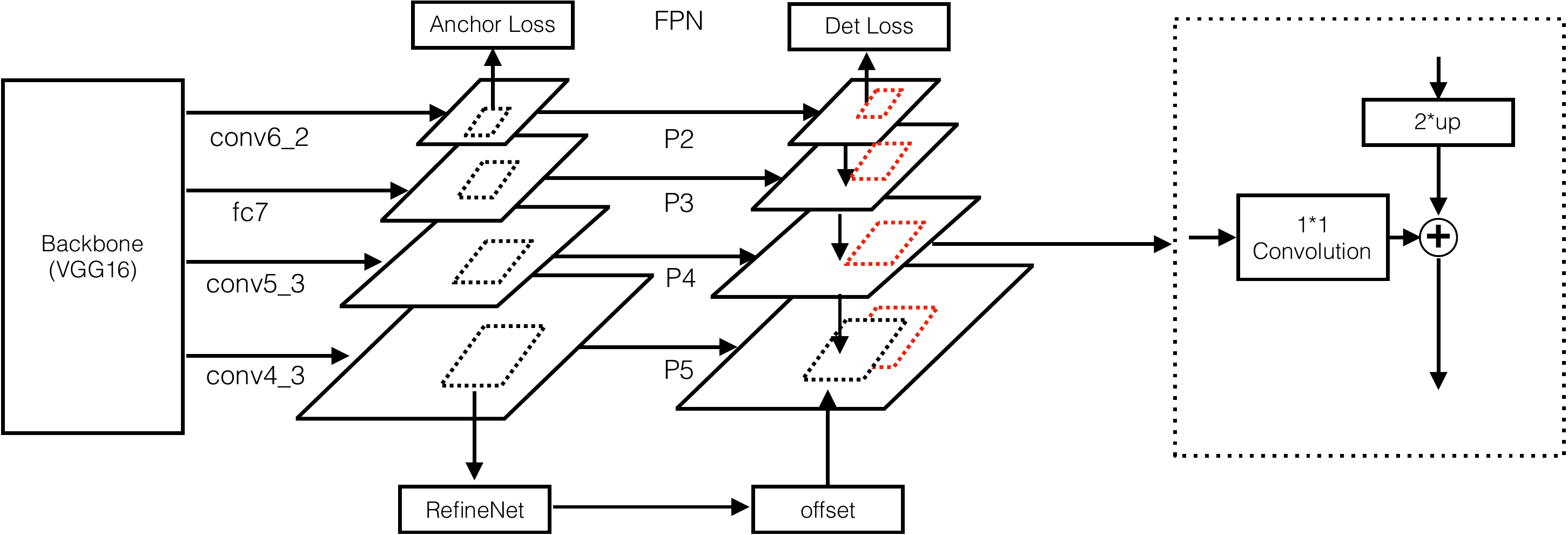}
\end{center}
   \vspace{-5mm}
   \caption{Framework of FPN with anchor refinement.}
\label{fig: Anchor Refinement}
\end{figure}

{\bf Backbone:} The backbone network structures of Brand Net are VGG-16\cite{simonyan2014very_VGG},ResNet50,ResNet101\cite{he2016deep_ResNet} and ResNeXt152, which are pre-trained on the ImageNet\cite{russakovsky2015imagenet_ImageNet}. Context feature fusion and resolution enhancement are keys to find small instances. In our network, feature pyramid network\cite{lin2017feature_FPN} and anchor refinement modual are joined in Brand Net to detect tiny brands. As shown in Fig.\ref{fig: Anchor Refinement}, Brand Net whose backbone is VGG-16 extracts features from layers: “conv4\_3”, “conv5\_3”, “fc7”, “conv6\_2” in a bottom-up pathway and “P2”, “P3”, “P4”, “P5” in FPN in a top-down pathway. It is a highly useful module for tiny brands proposals.

{\bf Anchor Refinement:} Anchor refinement network aims to remove negative anchors to reduce search area for the classifier and also coarsely adjust the locations and sizes of anchors to provide better initialization for the subsequent regression. Anchors after refinement with low confidence will be filtered. Then the number of anchors can be diminished from thousands to hundreds. Compared with fixed setting of anchor size when initializing network, it is a more effective anchors strategy for objects in a large scale. Fig.\ref{fig: Anchor Refinement} shows the anchor refinement used in FPN network. Anchor loss is similar to detection loss which combines classification loss and regression loss. Different from detection loss, anchor refinement is class-agnostic. The binary classification loss is the cross-entropy loss over two classes (foreground vs. background). Smooth L1 loss is used as regression loss.

\begin{figure}
\begin{center}
   \includegraphics[width=0.9\linewidth]{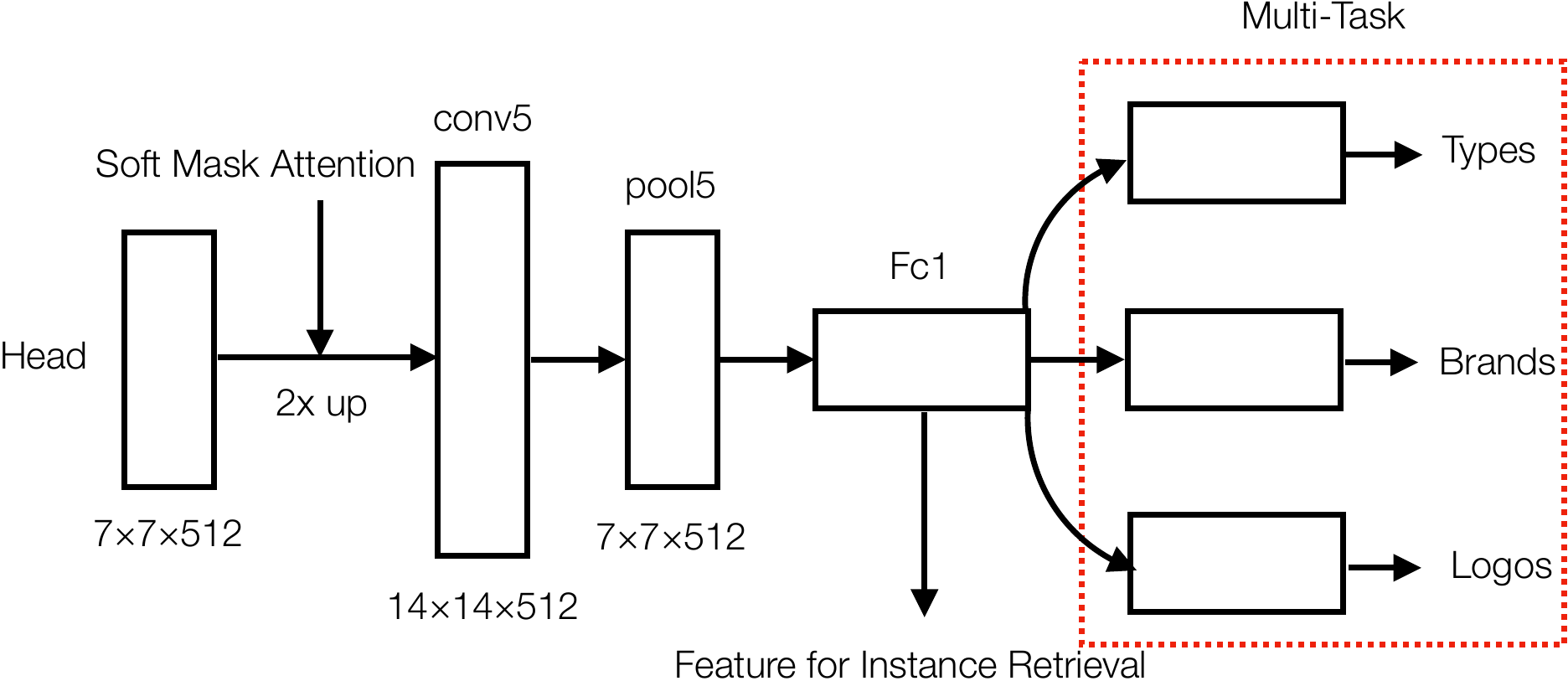}
\end{center}
   \vspace{-5mm}
   \caption{Framework of feature representation.}
\label{fig: Feature Representation}
\end{figure}

{\bf Feature Representation:} As shown in Fig.\ref{fig: Feature Representation}. The network structure of feature representation consists of convolution layers and fully connected layers, which is carefully designed for brand recognition. Multi-task training is used to optimize the embedding network. Finally, “Fc1” layer is extracted for instances retrieval.

\begin{figure}
\begin{center}
   \includegraphics[width=0.9\linewidth]{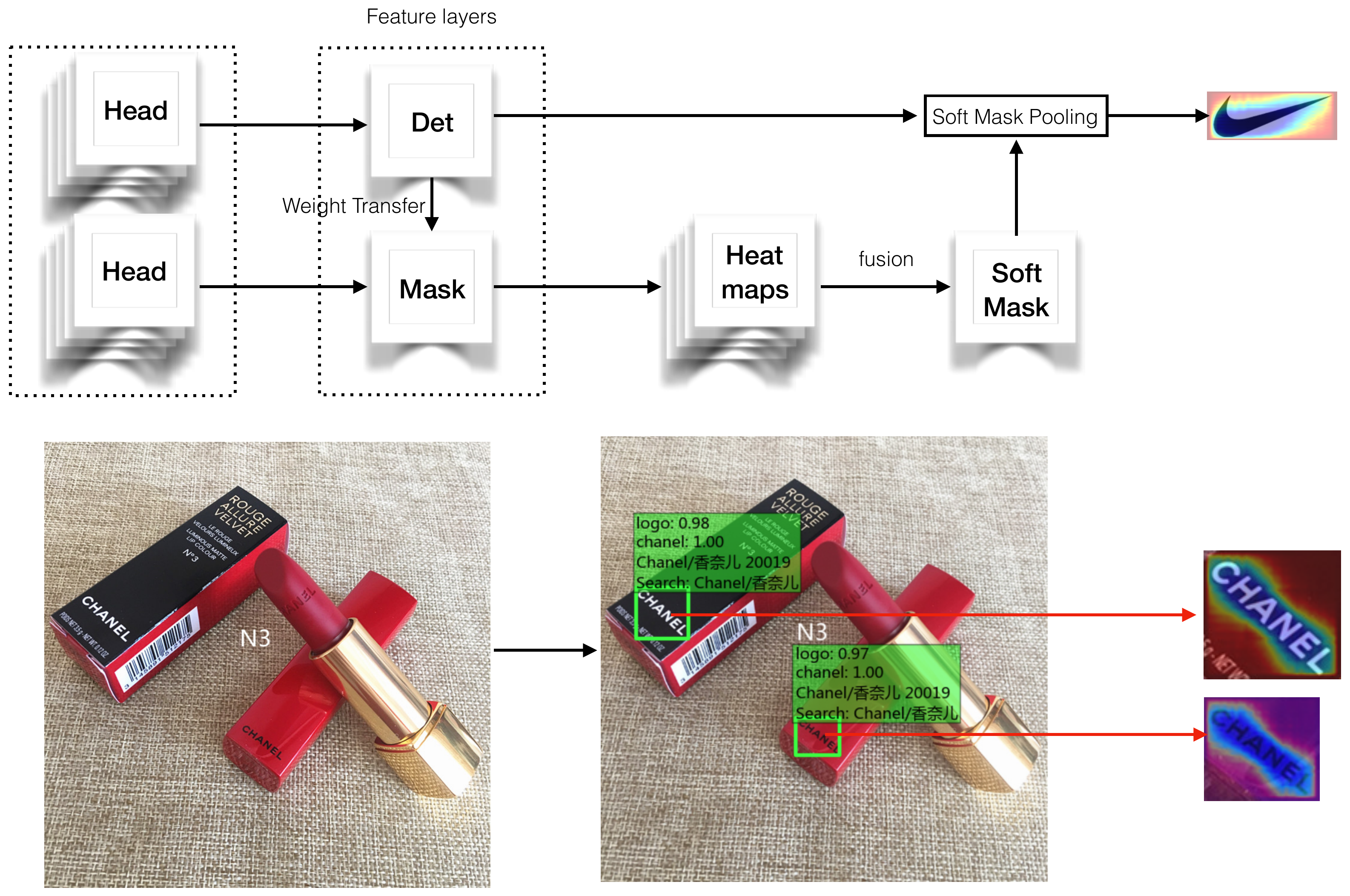}
\end{center}
   \vspace{-5mm}
   \caption{Framework of soft mask attention. Visualization is conducted in orginal image region with soft mask heatmaps.}
\label{fig: Soft Mask Attention}
\end{figure}

{\bf Soft Mask Attention:} Features of RoI extracted from bounding box regression will be affected by rotation, transformation and occlusion of brands, while features of RoI extracted from segmentation are more accurate for instances representation. In Brand Net, a branch for segmentation is added. Instead of predicting the hard masks from the original image coordinate space, class-agnostic soft masks of size $R\times K\times M\times M$ are directly utilized to form the attention heatmaps. $R$ represents the number of regions from RoIAlign layer.  $K$ represents the number of categories. It is 1 in our class-agnostic brands proposal network. $M = 28$ represents the size of heatmaps. Fig.\ref{fig: Soft Mask Attention} shows an example in which soft mask attention is applied in the brands representation in condition of rotation. The heatmaps obtained from soft mask can represent feature of brands more accurately.

{\bf Weight Transfer:} The annotations in Open Brands have only bounding boxes and categories, which lack pixel-to-pixel annotations. Some weakly supervised algorithm\cite{hu2017learning_WeightTransfer} can train segmentation with annotated bounding boxes. Supposed that category-specific coefficients \(w_{cls}^c\) represent weights of classification, \(w_{det}^c\) represent weights of regression and \(w_{seg}^c\) represent weights of segmentation. For a given category c, the weight transformation is a function which computes the relationship between \([w_{cls}^c w_{det}^c]\) and \(w_{seg}^c\).
\begin{equation}
\begin{split}
w_{seg}^c = f\left ( [w_{cls}^c w_{det}^c] | \theta \right)
\end{split}
\end{equation}
In Brand Net, $f\left ( \cdot \right )$ can be implemented as a small fully connected neural network, the learned parameters $\theta$ in the weight transfer function are class-agnostic. Therefore, the parameters can be applied in categories that are absent in the network.

{\bf Instances Retrieval:} For instance retrieval, the input is a 4096-dims vector calculated from feature embedding layers 'Fc1'. We use product quantization(PQ)\cite{jegou2011product_PQ}\cite{ge2014optimized_OPQ} to do a lossy compression for high-dimensional vectors, which allows relatively accurate reconstructions and distance computations in the compressed domain. Then inverted file (IVF)\cite{sivic2003video_IVF} is used for non-exhaustive searching. 

{\bf Loss and Training:} The entire training process is divided into 3 stages. At stage 1, we train an brand proposal network: 
\begin{equation}
\begin{split}
L_{step1} = L_{rpn} + L_{ar} + L_{det} 
\end{split}
\end{equation}
where $L_{rpn}$ is region proposal loss and $L_{ar}$ is anchor refinement loss taken as the class-agnostic smooth L1 loss and the binary cross-entropy loss. The class-aware smooth L1 loss and cross-entropy loss are utilized as the detection loss $L_{det}$. After that, we train segmentation branch and the weight transfer $f\left ( \cdot \right )$ on MSCOCO:
\begin{equation}
\begin{split}
L_{setp2} = L_{det} + L_{mask}
\end{split}
\end{equation}
we use the average binary cross-entropy loss as mask loss for training soft mask attention branch and weight transformation function $L_{mask}$. Finally, we train Brand Net end-to-end with fixed $f\left ( \cdot \right )$:
\begin{equation}
\begin{split}
L_{step3} = L_{rpn} + L_{ar} + L_{det} + L_{mt}
\end{split}
\end{equation}
$L_{mt}$ is a weighted cross-entropy loss used in embedding network. 
\begin{table}
\begin{center}
\resizebox{\linewidth}{!}{
\begin{tabular}{l c c c c}
\hline
Methods & Backbone & Resolution & mAP(0.5:0.95) & FPS \\
\hline
SSD & VGG & (512,512) & 31.8 & 22.5 \\
RefineDet & VGG & (512,512) & 40.2 & 26.7 \\
RetinaNet & ResNet50 & (800,1333) & 41.5 & 5.3 \\
RetinaNet & ResNet101 & (800,1333) & 45.7 & 4.7 \\
\hline
Faster R-CNN & ResNet50 & (800,1333) & 52.3 & 8.7 \\
Faster R-CNN & ResNet101 & (800,1333) & 56.4 & 4.7 \\
Faster R-CNN & ResNeXt152-32x8d & (800,1333) & 62.9 & 3.5 \\
\hline
Brand Net & VGG & (512,512) & 50.1 & {\bf 32.8} \\
Brand Net & ResNet101 & (800,1333) & 56.7 & 8.7 \\
Brand Net(SMA) & ResNet101 & (800,1333) & {\bf 60.4} & 8.3 \\
Brand Net(SMA) & ResNeXt152-32x8d & (800,1333) & {\bf 66.4} & 6.2 \\
\hline
\end{tabular}}
\end{center}
\vspace{-5mm}
\caption{Evaluation of Brand Net compared with state-of-the-art detection model on Open Brands. SMA is short for Soft Mask Attention.}
\label{tab: mAP of Final Experiments}
\end{table}

\section{results}
\label{sec:results}

We compare Brand Net with existing state-of-the-art one-stage methods SSD, RetinaNet, RefineDet and two-stage methods Faster R-CNN. RetinaNet is implemented jointly with FPN module. Faster R-CNN is implemented jointly with FPN module and RoIAlign. Different backbone networks and resolutions are taken into consideration.

{\bf Data:} Experiments are conducted on Open Brands. The test datasets of Open Brands contains nearly 60,800 images that are evenly sampled in the 1216 brands, even the smallest training category has only 157 images. This sampling strategy avoids the impact of category imbalances on the performance of mean average precision.

{\bf Detail:} For fair comparison, all models use synchronized SGD over 8 GPUs with a total with 16 images per batch. All models are trained for 270k iterations with an initial learning rate of 0.02, cosine learing rate strategy is used. Weight decay of 0.0001 and momentum of 0.9 are used. Considering that there are a lot of images without any instances in the real scene, we set threshold of confidence scores for evaluation to 0.5.

{\bf Speed:} We test the speed of the network inference in NVIDIA Tesla P100 GPU. The batch size is 1. As shown in Tab.\ref{tab: mAP of Final Experiments}, the frame per second (FPS) of Brand Net is higher than SSD (22.5) and RefineDet  (26.7) under the condition that VGG is adopted as backbone and the resolution is (512,512). Brand Net gets higher FPS and mAP than those in Faster R-CNN under the condition that the resolution are resized such that their scales (shorter edge) are 800 pixels and ResNet50, ResNet101 and ResNeXt152-32x8d are adopted as backbone. A detection network with large-scale categories need too many filters for class-aware layers which will extremely slow in practical applications. Brand Net gets a relatively nice performance in an efficient mode.

{\bf Performance:} We evaluate the performance with COCO metric: mean average precision (mAP). As shown in Tab\ref{tab: mAP of Final Experiments}, Brand Net(VGG) gets higher mAP than SSD(VGG) and RefineDet(VGG), it is even better than RetinaNet(ResNet101). Brand Net gets 0.03 higher mAP than Faster R-CNN(ResNet101) and gets 60.4(+3.7) mAP by jointing with soft mask attention module. The mAP of Brand Net(ResNet101) without SMA is nearly the same as the mAP of Faster RCNN(ResNet101), because the benifits of mAP brought by anchor refinement is equal to  those caused by region proposal network. We also test Faster R-CNN(ResNeXt152) on Open Brands. Brand Net gets 66.4 mAP which is 3.5 higher than Faster R-CNN(ResNeXt152). In summary, Brand Net that consists of region proposal, feature embedding and instance retrieval performs better on Open Brands. Deeper backbone, more context (FPN), better anchors (anchor refinement), more accurate representation of feature (SMA) all contribute to the improved mAP on Open Brands.

\section{conclusions}
\label{sec:conclusions}

This work presents Open Brands, a large-scale brand dataset with comprehensive annotations. Open Brands contains over 1,400,000 images and over 3,000,000 instances, which are richly labeled with fine-grained categories and bounding boxes. It surpasses existing logo datasets in terms of scale as well as richness of annotation. Compared with generic object detection, brand detection and recognition is a tough task due to tiny objects, large-scale categories, scalable network, similar brands and rich diversity. Brand Net learns brands features by jointly predicting locations and categories of brands. The estimated soft masks are used to pool or gate the learned features that lead to robust and discriminative representations for brands. Through extensive experiments compared with famous generic object detection network: SSD, RefineDet, RetinaNet, and Faster R-CNN, we demonstrate the effectiveness of Brand Net and the usefulness of Open Brands, which may significantly facilitate future researches.

\bibliographystyle{IEEEbib}
\end{document}